\documentclass[journal]{IEEEtran}
\usepackage{graphicx}
\usepackage{multirow}
\usepackage{algorithm}
\usepackage[noend]{algpseudocode}
\usepackage{subfig}
\usepackage{slashbox}
\usepackage{array}
\newcolumntype{P}[1]{>{\centering\arraybackslash}p{#1}}
\newcolumntype{M}[1]{>{\centering\arraybackslash}m{#1}}

\begin{document}
%
\title{Aerial~Map-Based~Navigation~Using~Semantic Segmentation~and~Pattern~Matching}
%
%
%

\author{Youngjoo~Kim \thanks{Youngjoo Kim is with Nearthlab, Seoul, South Korea (e-mail: rhymesg@gmail.com). This research has been conducted independently.}}

\maketitle

\begin{abstract}
This paper proposes a novel approach to map-based navigation system for unmanned aircraft. The proposed system attempts label-to-label matching, not image-to-image matching, between aerial images and a map database. The ground objects can be labelled by deep learning approaches and the configuration of the objects is used to find the corresponding location in the map database. The use of the deep learning technique as a tool for extracting high-level features reduces the image-based localization problem to a pattern matching problem. This paper proposes a pattern matching algorithm that does not require altitude information or a camera model to estimate the absolute horizontal position. The feasibility analysis with simulated images shows the proposed map-based navigation can be realized with the proposed pattern matching algorithm and it is able to provide positions given the labelled objects.
\end{abstract}


%

\section{Introduction}
%
%
%
%
\IEEEPARstart{M}{}ost unmanned aircraft rely on the global navigation satellite system (GNSS) to ascertain their position and velocity during flight. The GNSS is usually used to recalibrate the inertial navigation system (INS) periodically as the INS estimates a vehicle’s current position and velocity by integrating acceleration data over time. However, especially in military applications, various factors like jamming, interference, and unintentional interference due to terrain geometry can cause the GNSS device to work in an erroneous mode or shut down. Many researchers have been working in alternative/supplementary systems to cope with GNSS-devoid environments. These systems are required to control error accumulation of the INS and provide the absolute position in
terms of latitude and longitude.

One alternative is terrain-aided navigation (TAN) or terrain-referenced navigation (TRN) where the positioning data is provided by comparing the radar altimeter measurements of terrain elevation under an aircraft with an onboard digital elevation model (DEM). TAN has been considered as a solid alternative since it is all-weather capable in a sufficiently low flight altitude above terrain by utilizing a radar altimeter. However, the TAN system can be used in limited applications where an expensive, heavy altimetry sensor is available, e.g., cruise missiles or large aircraft. Moreover, it is well known that the TAN system mainly suffer from terrain ambiguity \cite{kim2017utilizing} and slant range measurements \cite{spiegel2016slant} due to the nature of one-dimensional, wide-lobe radar altimeter measurements.

Another alternative is vision-based navigation where images taken by a camera on an aircraft are used to estimate the position of the aircraft. There are two features that make this approach attractive; first, cameras are passive sensors, so it is hard to detect or interfere with them. Second, because most unmanned aircraft are already equipped with cameras, they don’t have to mount additional payload for utilizing vision data to navigate. Earlier works on vision-based navigation addressed positioning by obtaining elevation data from aerial images and then matching it to a DEM \cite{sim2002integrated, rodriguez1990matching}. These approaches can be viewed as a two-dimensional extension of TAN. A recently proposed work uses a stereo analysis of the image sequence to obtain heights of the feature points and compare them with the DEM to estimate the vehicle state \cite{kim2018vision}. However, the so-called vision-based TAN approach is limited because the performance depends on the resolution and accuracy of the DEM. Furthermore, in practice the visual terrain surface elevation can differ from the DEM that is usually obtained by a synthetic aperture radar (SAR). Hence, careful construction of the digital surface model (DSM) as in \cite{el2014digital} is required to acquire an elevation model of the visual terrain surface.

Map-based navigation approaches have been attracting attention since recently public map databases that render 2D locations on aerial/satellite imagery have become available, including Google Maps, Airbus Defence and Space, and OpenStreetMap. Thus, once established, the map-based navigation technology will require no map-building process and can be scaled to various types of aircraft systems and map databases. Several approaches have been attempted to match images taken by an aircraft-mounted camera to the public map: image registration by adopting a correlation filter \cite{shan2015google} and feature point detection and matching between two scenes \cite{zhuo2017automatic}. Mountain drainage patterns \cite{wang2016characterization} and road intersection \cite{volkova2018more} have also been used to characterize scenes. However, relevant works have reported that variations in scale, orientation, and illumination pose challenges to these vision-based approaches. Moreover, because the imagery in a public map database is not regularly or consistently updated, the aerial images recorded in flight may differ due to seasonal changes. A more robust technique for scene matching is required for realizing a reliable map-based navigation system. The huge amount of storage required for the image database is another problem that threatens practicality of the map-based navigation.

To tackle the problems above, this paper proposes the novel use of deep learning technologies as a tool for extracting high-level features, called ``labels'', from aerial images and a map database. The deep learning is supposed to do what it does well as a part of an image processing system, rather than more complex tasks like image matching and navigation. For example, once ground objects such as road intersections, buildings, and highways are distinguished, not necessarily identified, the configuration of the objects can be used to find corresponding location in the map database. In other words, the aerial localization is done by pattern matching of labelled objects, not image-to-image matching. Semantic segmentation is known to provide accurate object position in the image. However, other methods like object detection can be also employed compromising accuracy for speed. In this way, the aerial images are converted to a set of dense information that is robust to imagery variations and noises, requiring significantly less storage and computational power. To show the feasibility of such an approach, this paper proposes and verifies a pattern matching algorithm for estimating the camera's position assuming the image processing is done.

The rest of this paper is organized as follows. It starts with addressing the proposed map-based navigation system in Section \ref{sec:system}. The method to utilize the labelled objects for pattern matching is discussed in detail. The feasibility the proposed approach with simulated images is discussed in Section \ref{sec:simulation}. Finally, Section \ref{sec:conclusion} gives summary and conclusion.

\begin{figure}[t!]
	\centering
	\includegraphics[clip,width=1.0\linewidth]{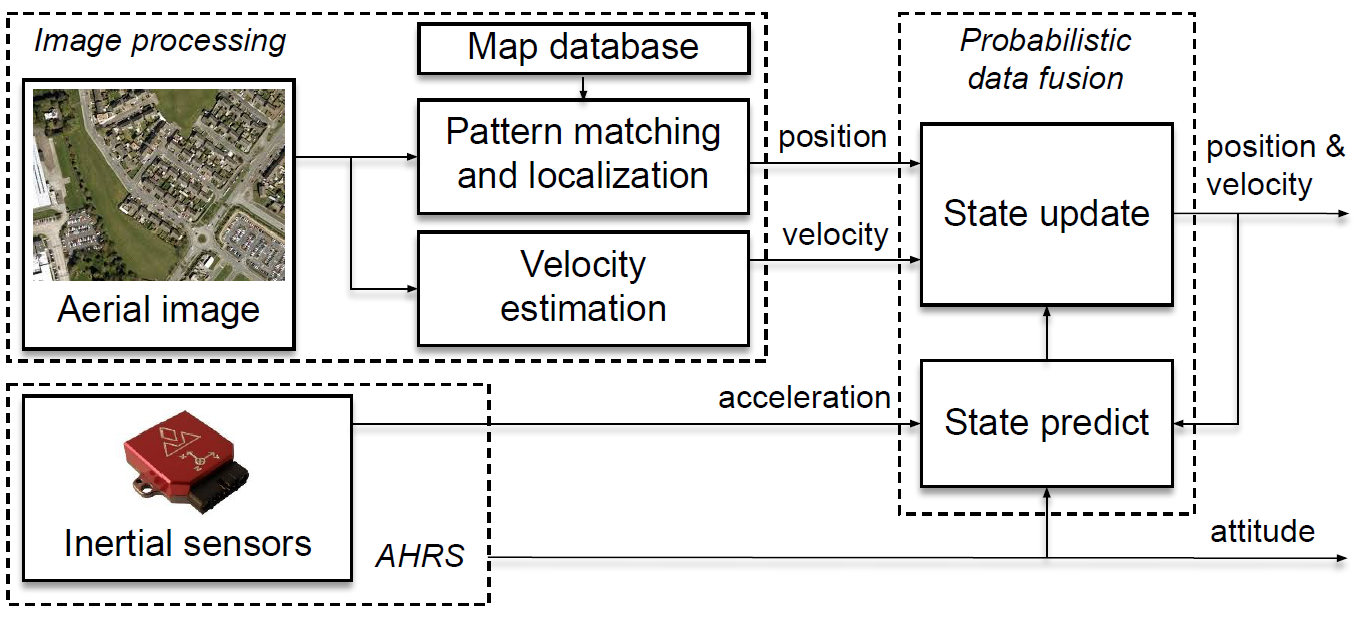}%
	\caption{Block diagram of the proposed map-based navigation system.}
	\label{fig:diagram}
\end{figure}

\section{Proposed Map-Based Navigation System}\label{sec:system}
The block diagram of the proposed map-based navigation system is presented in Fig. 1. The overall system consists of image processing, probabilistic data fusion, and attitude and heading reference system (AHRS) blocks. Output of this system is the primary aircraft states: position, velocity, and attitude. The key idea of the proposed approach is depicted in Fig. 2, which consists of image processing and pattern matching. These are discussed in detail below in Section \ref{sec:semantic} and \ref{sec:pattern}, followed by brief remarks on velocity estimation and probabilistic data fusion.

\begin{figure}[t!]
	\centering
	\includegraphics[clip,width=0.95\linewidth]{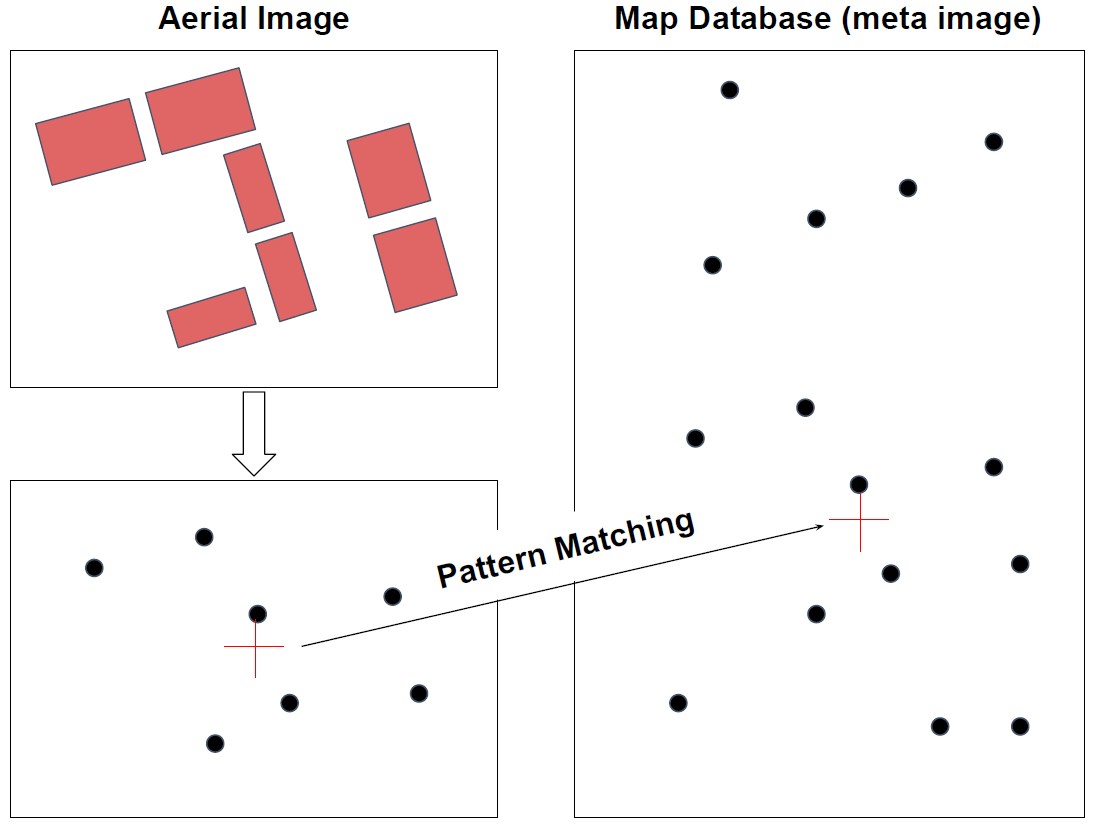}%
	\caption{The configuration of labelled ground objects is used to match the aerial scene with the meta image of the database. The red rectangles denote the result of semantic segmentation on an aerial image. The black dots denote the labelled objects. The red cross denotes the center of the image.}
	\label{fig:pattern}
\end{figure}

\subsection{Image Processing}\label{sec:semantic}
The navigation is started by the image processing block that classifies and localizes (within an image) various ground objects. Deep learning systems have been proven to very effective in separating the objects from the background and classify them. State-of-the-art approaches of deep learning can be used as a tool for object labelling. For example, Mask R-CNN \cite{he2017mask} has shown superior performance on segmentation and classification. Successful semantic segmentations on aerial images have also been reported \cite{marmanis2016semantic, mallberg2018map, niu2021hybrid}. Besides, the oriented object detection methods on aerial images \cite{xie2021orcnn, yang2021kl} are emerging as attractive candidates. The recent progress of oriented object detection is reporting evolving accuracy, featuring its inherent efficiency.

Each ground object can be designated a label such as building, stadium, park or green area, road intersection, lake or river, agricultural field, and mountain. These ground objects are very robust about preserving their shape. The objects in the image can be represented as dots by averaging the pixels occupying each object. Constructing such an on-board meta-image map should also be done where the objects in the map are classified in the same way for the aerial scene.

Note that the term ``localization'' in image processing or deep learning means obtaining objects' positions in the image, e.g., in pixels. On the other hand, the purpose of the overall navigation is to obtain position of the camera in the reference frame the map database is based on, e.g., in meters.

\subsection{Pattern Matching and Localization}\label{sec:pattern}
The configuration of the objects in the image is compared to the configuration of those in the map database. This is called pattern matching here, not landmark matching, because the random ground objects in a scene are used to match the scene against the map. In other words, it is not required to identify each ground object. In this way, the proposed approach lowers the burden of image processing and therefore it is expected to increase robustness of the vision-based navigation system. Since the objects are represented as dots, the pattern matching problem involves scale-invariant and rotation-invariant features.

\begin{algorithm*}[tb!]
	\caption{Proposed Pattern Matching Algorithm}
	\begin{algorithmic}[1]
		\State $N_{best} = 0$
		\State $E_{best} = \inf$
		\State $c_{best}^D = \textnormal{None}$
		\For {$(o_i^I,o_j^I) \in C(O^I)$}
		\State \textit{Calculate polar coordinates from the center of the image:} $(r_i, \theta_i), (r_j, \theta_j)$
		\State $\Delta \theta_j = \theta_i - \theta_j$
		\For {$(o_i^D,o_j^D) \in P(O^D)$ \textbf{if} $label(o_i^D) = label(o_i^I)$ \textbf{and} $label(o_j^D) = label(o_j^I)$}
		
		\State \textit{Find the origin of the polar coordinates on database,} $c_{ij}^D$\textit{, satisfying} \newline
		$~~~~~~~~~~~~\Delta \Theta_j = \Theta_i - \Theta_j = \Delta \theta_j$ \textit{and} $R_i : R_j = r_i : r_j$ \newline
		$~~~~~~~~~~~~$\textit{where} $(R_i, \Theta_i), (R_j, \Theta_j)$ \textit{are polar coordinates of} $o_i^D,o_j^D$ \textit{with origin} $c_{ij}^D$
		\State $N_{matched} \leftarrow 2$
		\For {$o_k^I \in O^I - \{o_i^I,o_j^I\}$}
		\State \textit{Calculate} $r_k, \theta_k$ \textit{and therefore} $r_k / r_i, \Delta \theta_k$
		\For {$o_k^D \in O^D - \{o_i^D,o_j^D\}$ \textbf{if} $label(o_k^D) = label(o_k^I)$}
		\State \textit{Calculate} $R_k, \Theta_k$ \textit{and therefore} $R_k / R_i, \Delta \Theta_k$
		\If {$|r_k / r_i - R_k / R_i| < \delta_r $ \textbf{and} $|\Delta \theta_k - \Delta \Theta_k|< \delta_{\theta}$}
		\State $e_k = |r_k / r_i - R_k / R_i| + |\Delta \theta_k - \Delta \Theta_k|$
		\State $N_{matched} \leftarrow N_{matched} + 1$

		\EndIf
		\EndFor
		\EndFor
		
		\If {$N_{matched} >= N_{min}$}
		\State $E_{ij} = std(\{e_k\})$
		\If {$N_{matched} >= N_{best}$ \textbf{and} $E_{ij} < E_{best}$}
		\State $N_{best} \leftarrow N_{matched}$
		\State $E_{best} \leftarrow E_{ij}$
		\State $c_{best}^D \leftarrow c_{ij}^D$
		
		\EndIf
		
		\EndIf
		
		\EndFor
		\EndFor
		
		\State \textit{Return} $c_{best}^D$
	\end{algorithmic}
\end{algorithm*}

Here a random sample consensus (RANSAC) \cite{fischler1981random} based method is proposed in Algorithm 1. The consensus algorithm will iteratively test hypotheses of matching between the labels in an image and those in the database to provide one or multiple matching candidates where each label in the image has the same relative angle and distance to the equivalent label in the database. Suppose objects in the image and the database are denoted as $o^I$ and $o^D$, respectively. Each object pertains to its label and 2-dimensional position in its own coordinate system. It is assumed the image is taken by a downward-looking camera. If the attitude of the camera is not zero, the attitude information can be incorporated to project the image objects onto the plane parallel to the ground before executing the algorithm. The set of objects in the image is denoted as $O_I$, and that in the database as $O^D$. For every 2-combination of the objects on the image, $(o_i^I,o_j^I) \in C(O^I)$ where $C(O^I)$ means all the 2-combinations of the image objects, the polar coordinates of the two objects are calculated. Taking the polar coordinates of the first object $o_i^I$ as the reference, the relative radius and angle, $r_j / r_i$ and $\theta_i - \theta_j$, are compared as depicted in Fig. 3. All the angle differences here are represented in $(-\pi,\pi]$. For every 2-permutation of objects in the database, $(o_i^D,o_j^D) \in P(O^D)$, find the origin of the polar coordinates that makes the same configuration that $(o_i^I,o_j^I)$ does to the image center. This can be done by finding the intersection of two circles with radius $R_i$ and $R_i \times r_j/r_i$, centered at $o_i^D$ and $o_j^D$, respectively. The intersection with the same sign of the relative angle is chosen out of at most two intersections. Every time the match between the image object and the database object with the same relative radius and angle with tolerances $\delta_r$ and $\delta_{\theta}$ is found, increment the number of matched points $N_{matched}$ and store the matching error $e_k$. The matching candidate with the lowest standard deviation of matching error is chosen as the best match and the pattern matching outputs the corresponding horizontal position. Note that the proposed algorithm doesn't require a camera model or altitude information to get the horizontal position. The coordinates and labels of the objects in the image and the database are the only input.

\begin{figure}[b!]
	\centering
	\includegraphics[clip,width=0.8\linewidth]{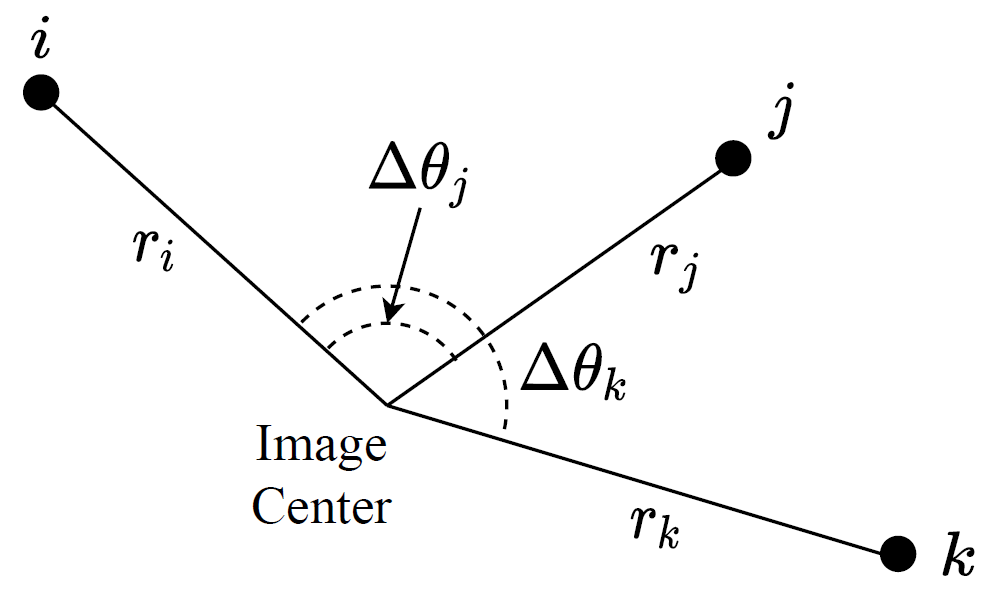}%
	\caption{The polar coordinates of the image objects with respect to the image center.}
	\label{fig:matching}
\end{figure}

Several variants of the algorithm are available. Algorithm 1 attempts to match the objects in the image to those in the whole database. In practice, one can guess a region the objects are probably within by using the results of the probabilistic data fusion technique, which is discussed in Section \ref{sec:filter}. Searching for objects in a smaller region will reduce the computation time and the number of false positives. Plus, whereas Algorithm 1 tests if the objects have the same label, classifying the objects into a smaller set of labels, or only one label, is possible to get more robust matches compromising uniqueness. If ambiguity arises and it is hard to determine the best match, it is able to output multiple candidates and let the probabilistic filter deal with them. Moreover, if the camera model is known, the height above ground can be induced from the projection of the matched data objects into the image. Users can exploit any variant design to meet their needs.

\subsection{Velocity Estimation}
The image processing block should provide both velocity and position as an alternative to GNSS for unmanned aircraft navigation. Although the observability in velocity is required to correct the INS drift, many existing studies haven’t addressed velocity estimation for vision-based navigation. It is well known that the velocity can be measured using optical flow techniques. Or one can borrow the insights from a recent work \cite{bloesch2017iterated} for visual-inertial odometry technique.

\subsection{Probabilistic Data Fusion}\label{sec:filter}
Designing a probabilistic filter requires analysis on the measurement and error models of the localization and velocity estimation blocks. It is obvious that an altimeter is required for the altitude information because the pattern matching and localization block provides horizontal position. This 3-dimensional position and the 3-dimensional velocity from the velocity estimation block will construct the measurement. If it turns out the measurement errors can be modelled as additive Gaussian noise, as in GNSS, borrowing a simple INS/GNSS filter \cite{kim2019introduction} will be desirable. If the pattern matching and localization block is designed to provide multiple candidates of aircraft position from a scene, possibly a particle filter dealing with a multimodal distribution will be suitable.

In cases where the navigation performance depends on aircraft states, information theory can be exploited to quantify and control the information gathered by the sensors. Background in information theory \cite{kim2019real} would be helpful in this problem. For example, desirable or undesirable trajectory can be analyzed to meet the criteria of the navigation performance.

The probabilistic filter provides the estimated state and its covariance as an output. The covariance is the measure of uncertainty of the estimate. The uncertainty information can be used to restrain the region of interest on the database to reduce false matchings computational burden in pattern matching.

\section{Feasibility Verification}\label{sec:simulation}
The feasibility of the proposed approach described above is verified by a simulation. Suppose it is able to get labelled objects of aerial images and a map database. It is tested if the proposed pattern matching and localization, Algorithm 1, works with the simulated image objects under various conditions.

\begin{figure}[t!]
	\centering
	\includegraphics[clip,width=0.95\linewidth]{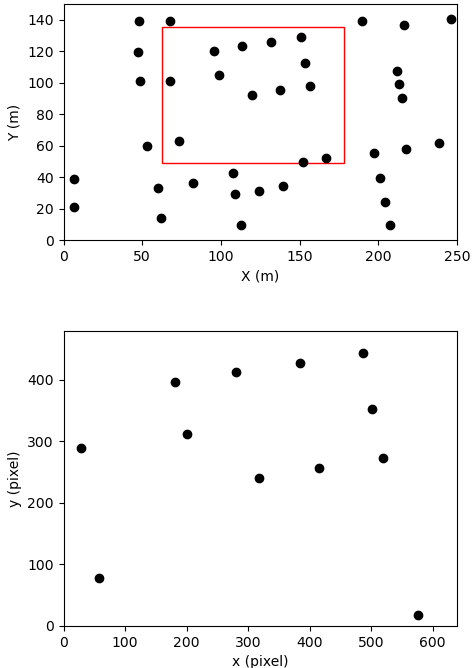}%
	\caption{The graph at the top represents positions of objects in the database used in the simulation. The red rectangle denotes an area the camera takes an aerial image for an instance. The resulting image is simulated as the graph at the bottom.}
	\label{fig:simulation}
\end{figure}

\subsection{Method}
Suppose a database meta image contains labelled objects and their absolute horizontal positions in meters. A unmanned aircraft with a downward-looking camera takes aerial images. The simulated images are generated by using the true absolute horizontal position and the true height above ground, the camera's field of view (FOV), the image size in pixels, and errors in camera attitude and image processing. For example, the squared region in Fig. 4 is projected onto the image on the right. The database represents objects with the same label in a region of 250$\times$150 meters. The image size is 640$\times$480 pixels.

\begin{table*}[t]
	\captionsetup{justification=centering}
	\caption{Control variables and result of the simulation.}
	\centering
	\label{tab:result}
	\renewcommand{\arraystretch}{1.5}
	\begin{tabular}{|c|l|M{2.0cm}|M{2.0cm}|M{2.0cm}|M{2.0cm}|M{2.0cm}|}
		\hline
		\multicolumn{1}{|l|}{}      &                          & Case 1      & Case 2         & Case 3           & Case 4         & Case 5        \\ \hline
		\multirow{3}{*}{Variables} & Attitude error std (deg) & 0           & 0.05           & 0.05             & 0.15           & 0.15          \\ \cline{2-7} 
		& Pixel error std (pixel)  & 0           & 1              & 3                & 3              & 3             \\ \cline{2-7} 
		& Field of view (deg)      & 35          & 35             & 35               & 35             & 45            \\ \hline
		\multirow{3}{*}{Result}     & Error std (m)            & 0.01        & 0.53           & 1.97             & 1.74           & 3.29          \\ \cline{2-7} 
		& \# of rejected images       & 0/500 (0\%) & 22/500 (4.4\%) & 107/500 (21.4\%) & 95/500 (19\%)  & 0/500 (0\%)   \\ \cline{2-7} 
		& \# of false positives       & 0/500 (0\%) & 2/478 (0.4\%)  & 30/393 (7.6\%)   & 36/405 (8.9\%) & 9/500 (1.8\%) \\ \hline
	\end{tabular}%
	\renewcommand{\arraystretch}{1}
\end{table*}

The attitude error and image processing error are also reflected. The attitude error is contributed by the gimbal control error. The pattern matching algorithm assumes the camera is looking straight down, but actually the camera is misaligned by the attitude error in the simulated image. Note that the algorithm is not affected by yaw angle (horizontal rotation) or its error. The database and the image in the simulation are aligned in yaw axis, for better presentation without loss of generality, and the errors in pitch and roll are used. And the imperfect image processing causes the pixel error on x, y coordinates of each object in the image. All the errors are sampled from normal distributions of given standard deviations.

Using the database meta image in Fig. 4, 500 random true positions from uniform distributions in x, y axes are sampled. The set of objects in each image and the set of objects in the map database are fed into Algorithm 1. The result of Algorithm 1 on the simulated images is compared to the true positions. The images are taken at $100 m$ above the ground, and the FOV and error statistics are controlled to see the effect of the number of objects in the image and the type of errors. The attitude error of $0.05^{\circ}$, slightly larger than that of an industry level AHRS, and the pixel error of $1$, as the discretization error, are taken as the reference error statistics. By increasing each threefold, the effect of each error is observed. FOVs of $35^{\circ}$ and $45^{\circ}$ are tested to see the effect of the number of objects in the images. All the cases are run with the same parameters: $N_{min} = 6$, $\delta_{\theta} = 0.2$, $\delta_{r} = 0.2$. Larger threshold values tends to accept more false positives and smaller threshold values tends to reject more matches.

\subsection{Result}
Table 1 shows the result of the simulation on 5 different cases. The standard deviation of the horizontal distance error are presented with the number of rejected matching instances and the number of false positive matchings. Cases 1-4 are with the FOV of $35^{\circ}$ where the average number of objects taken in the images is about 17. Case 1 is the result with no attitude error and no pixel error. It shows an almost perfect matching and the nonzero position estimation error comes from the generous thresholds in the matching algorithm. Case 2 is the result with the attitude error of $0.05^{\circ}$ and the pixel error of $1$. One can observe the pattern matching algorithm is able to estimate the absolute position with reasonable estimation errors. $4.4 \%$ of matchings are rejected in this case. This happens when the position estimate is not provided because the number of matched points are less than $N_{min}$. The false positive is the case when the algorithm finds a wrong matching. The large number of false positives is observed when the pixel error is increased to 3 as in Case 3. This can be mitigated by using a prior knowledge of the region the camera might be taking to reject unreasonable outputs. The attitude error is increased in Case 4. It doesn't necessarily affect the matching itself because the relative configuration of the image objects is not changed by the attitude error. The estimation error as well shows no noticeable difference from Case 3. The effect of the attitude error might be relatively smaller than the pixel error where $0.15^{\circ}$ of angle error corresponds to $0.26 m$ on the ground. The FOV is increased to $45^{\circ} $ in Case 5 where the average number of objects in the images is about 34. In this case, no matching is rejected and only $1.8 \%$ of the matchings are false positives despite the large pixel error. Capturing more objects helps finding the match, but the decreased resolution might cause larger position estimation error.

\section{Conclusion}\label{sec:conclusion}

The proposed method aims for realizing a robust map-based navigation system for unmanned aircraft. In summary, the pattern matching and localization block compares the aerial images with a map database in a robust, efficient way to provide the horizontal position. The key idea is to let the map matching algorithm deal with the high-abstraction information of the image, rather than the image itself. The map-based navigation system is expected to continuously provide the position and velocity by probabilistic data fusion of the position information from the pattern matching with the velocity from visual-inertial odometry.

The feasibility analysis with simulated meta images shows that the proposed pattern matching algorithm can provide position estimates by using labelled objects on the images and the database. The proposed approach to map-based navigation would be an attractive choice for image-based navigation if the image processing block is able to provide labelled objects and their positions in the image.


%

%

\ifCLASSOPTIONcaptionsoff
  \newpage
\fi



%
%
%

\bibliographystyle{IEEEtran}
\bibliography{map_based_bib}

%





\end{document}